# Informal Persian Universal Dependency Treebank

**Roya Kabiri, Simin Karimi, Mihai Surdeanu**

University of Arizona
Tucson AZ, USA
{royakabiri, karimi, msurdeanu}@email.arizona.edu

**Abstract**
This paper presents the phonological, morphological, and syntactic distinctions between formal and informal Persian, showing that these two variants have fundamental differences that cannot be attributed solely to pronunciation discrepancies. Given that informal Persian exhibits particular characteristics, any computational model trained on formal Persian is unlikely to transfer well to informal Persian, necessitating the creation of dedicated treebanks for this variety. We thus detail the development of the open-source Informal Persian Universal Dependency Treebank, a new treebank annotated within the Universal Dependencies scheme. We then investigate the parsing of informal Persian by training two dependency parsers on existing formal treebanks and evaluating them on out-of-domain data, i.e. the development set of our informal treebank. Our results show that parsers experience a substantial performance drop when we move across the two domains, as they face more unknown tokens and structures and fail to generalize well. Furthermore, the dependency relations whose performance deteriorates the most represent the unique properties of the informal variant. The ultimate goal of this study that demonstrates a broader impact is to provide a stepping-stone to reveal the significance of informal variants of languages, which have been widely overlooked in natural language processing tools across languages.

## 1. Introduction

Most studies in natural language processing in general and dependency parsing in particular have mainly focused on the formal variants of languages and disregarded the informal ones (if any), with the notable exception of Twitter parsing and related tasks (Kong et al., 2014; Bharti et al., 2015).

Persian, a southwestern Iranian language spoken in Iran[1], has two distinct variants: the literary formal form that has been traditionally used in writing as it was regarded more prestigious and prescriptively accurate, and the conversational informal form typically used in oral communications.[2] The phonological, morphological, and syntactic distinctions between the two variants reveal that these varieties have fundamental differences that cannot be attributed solely to pronunciation discrepancies, and that they are more like two closely related dialects than simply different styles, which is the case in some languages such as English. Several studies have suggested that contemporary Persian is diglossic (Boyle, 1952; Jeremiás, 1984), a term first coined by Ferguson (1959) to describe a speech community that naturally uses two varieties of a language for different purposes. With the advent of social media, the restrictions against the use of the conversational variant in writing have been challenged, and speakers tend to write in the spoken informal variety rather than the written formal variety, notwithstanding the strong opposition from the academy of Persian language and literature. Using conversational Persian in the Web creates a new challenge for the analysis of informal texts, as current grammars, academic textbooks, corpora, and computational models of this language focus mainly on the formal variant.

In this study, we show that the unique characteristics of informal Persian and the lack of an informal Persian annotated corpus necessitate the development of a dedicated treebank for this variant. We then describe the Informal Persian Universal Dependency Treebank (iPerUDT), annotated in CoNLL-U format[3] within the Universal Dependencies (UD) scheme. The treebank is released under an open-source license.[4] We also show that two parsers, primarily trained on formal data, face more unknown tokens and structures when evaluated on informal data, and fail to generalize to the basic patterns, thus making an ill prediction in such conditions. Finally, we demonstrate that the most erroneous tokens as well as the dependency relations whose performance deteriorates the most represent the unique properties of informal Persian.

## 2. Related Work

The two existing Persian UD treebanks have been primarily concerned with the formal literary variant: Uppsala Persian Universal Dependency Treebank (Uppsala UDT) (Seraji et al., 2016) includes formal texts found in fiction, news reports, technical descriptions, and cultural texts; and the Persian Universal Dependency Treebank (PerUDT) (Rasooli et al., 2020), automatically converted from the Persian Dependency Treebank (Rasooli et. al 2013) to UD, also contains genres such as news, academic dissertations, fiction, and travel reports. Rasooli and colleagues (2013) note that the presence of colloquial writings makes computational processing of Persian texts very challenging. When creating the Persian Dependency Treebank, they manually removed sentences containing colloquial words (Rasooli et. al. 2013: 312). The Uppsala UDT includes only a few informal sentences, however, they are mostly annotated incorrectly or incompletely. For instance, *del-am-o* in (1) is annotated as a single token, with no lemma in this treebank although it is a multiword expression made up of the noun *del* 'heart', the 1SG possessive pronoun -*am* 'my' and the differential object marking -*o*. Note that

---

[1] Persian is spoken in Iran (Farsi), Afghanistan (Dari) and Tajikestan (Tajiki). The dialect we examine in this work is a variant of Farsi, called Tehrani, the standard colloquial dialect spoken in Iran. We utilize the term Persian to refer to this variant throughout this work.

[2] We alternate formal and informal language terms with 'literary/written grammar or style' and 'colloquial/spoken/conversational grammar or style', respectively, in this study. All of these terms refer to the same entity.

[3] https://universaldependencies.org/format.html

[4] https://github.com/royakabiri/iPerUDT

Persian features the use of differential object marking to morphologically mark specific objects (see Karimi and Smith, 2020 for details). This marker appears on definite objects as well as specific indefinite objects regardless of animacy.[5]

(1) del-am-o       xun    na-kon
    heart-1SG-DOM  blood  NEG-do
    'Don't upset me!' (lit. 'Don't make my heart bloody!')

Furthermore, dependency parsers for Persian (Seraji et al., 2012; Khallash et al., 2013; Seraji et al, 2015) have been developed for the formal variant due to the availability of the annotated formal datasets. However, there has been no work on parsing the informal variant. We experiment with two open-source dependency parsing methods that achieved state-of-the-art performance for English, Stanza (Qi et al., 2020) and PaT (Vacareanu et al., 2020) to investigate the parsing of informal Persian. Although both parsers rely on a bidirectional LSTM (BiLSTM), the former is a graph-based neural dependency parser, and the latter is a sequence model, treating dependency parsing as a sequence tagging problem. We show that the architectural differences between the two parsers play a significant role in their performance differences, as PaT requires a large amount of training data to converge, whereas Stanza does not.

## 3. Informal Persian Treebank

### 3.1 Formal and Informal Persian Distinctions

Persian has twenty-nine phonemes, including twenty-three consonants and six vowels, conveyed by thirty-two letters of the Persian alphabet (University of Victoria Phonetic Database, 1999). This language is written from right to left in Perso-Arabic script, an extended version of the Arabic script. Vowels are categorized into two groups in terms of length: short vowels (/e o a/) and long vowels (/i u â/). The former is represented by superscript or subscript diacritics attached to the letters of the alphabet while the latter is represented by the letters of the alphabet. However, in practice, short vowels are not inscribed in the language. In general, Persian orthography causes problems when it comes to text processing due to its characteristics, including but not limited to phoneme diversity (i.e. representing a phoneme by different letters), unwritten short vowel diacritics, lack of capitalization, and the optionality to switch between a variety of space characters with different widths (whitespace, zero width non-joiner, and no space) (see Seraji, 2015 for details).

Persian is a null subject language which also exhibits rich inflected verbal morphology (see Karimi, 2005). Persian grammar features the use of complex predicate constructions, which are verbal structures consisting of more than one word that convey information that is normally expressed by a single simple verb in a language like English. Persian complex predicates consist of a non-verbal element (NVE) and a light verb (LV); for instance, *tamiz kardan* lit. 'clean doing', means 'to clean'. Informal Persian possesses a wide range of phonological, morphological, and syntactic properties that are different from formal Persian. Due to space constraints, we will briefly discuss only certain distinctive characteristics of this variant in what follows.

#### 3.1.1 Phonology

Since the writing in blogs and social media reflects the way people speak, changes in pronunciation are represented in such texts, resulting in different word forms in informal Persian compared to the formal version of this language. In informal Persian, when the syllable-final alveolar nasal consonant /n/ is followed by the voiced bilabial stop /b/, it assimilates to the bilabial nasal /m/ (*panbe→pambe* 'cotton', *shanbe→shambe* 'Saturday'). This is known as anticipatory assimilation, because the gesture for one sound is affected by anticipating the gesture for the following sound (Ladefoged and Johnson, 2011). Similarly, the syllable-initial voiceless alveolar stop /t/ fully assimilates to its preceding voiceless alveolar fricative /s/ within a word or at word boundaries to adopt its manner of articulation feature (*xaste→xasse* 'tired', *xâst-am→xâss-am* 'I wanted'). Informal Persian also exhibits vowel harmony in words with CVCVC syllable structure by raising the mid vowels /e/ and /o/ to harmonize with the following corresponding high vowels /i/ and /u/ (*belit→bilit* 'ticket, *sholugh→shulugh* 'crowded'). Moreover, although Persian phonotactics features syllable-final consonant clusters of up to two consonants, it does not allow initial consonant clusters. Nonetheless, the syllable-final clusters are rarely pronounced in the informal variant in that the second consonant in a sequence of two consonants is typically elided (*dust→dus* 'friend', *fekr→fek* 'thought'). However, there are several cases to which consonant cluster simplification does not apply (*shokr→\*shok* 'praise', *kabk→\*kab* 'partridge' (The asterisk denotes ungrammaticality)). Furthermore, the vowel /â/ in formal Persian alternates to /u/ before nasal consonants either intra-syllabic or inter-syllabic in informal Persian (*bârân→bârun* 'rain', *xâne→xune* 'house'). Despite that this phonological rule appears to be systematic at first glance, the alternations contain numerous irregularities (*alân→\*alun* 'now', *kârgardân→\*kârgardun* 'director'). Although many phonological differences between formal and informal Persian are predictable due to the phonological rules discussed, there are several others that are unpredictable.

#### 3.1.2 Morphology

The plural marker in Persian noun morphology is the suffix *-hâ*. However, *ketâb-hâ* 'books' would be pronounced and written as *ketâb-â* in the informal variant. If the plural marker follows a vowel, it will be realized as *-jâ* to avoid hiatus (*zendegi-hâ→zendegi-jâ* 'lives'). Within the realm of verbal morphology, the agreement inflections appear in slightly different forms than in formal Persian (e.g., 3SG: *-ad* vs *-e*; 2PL: *-id* vs *-in*; 3PL: *-and* vs *-an*). The verbal construction has also undergone a dramatic change. While the copula *budan* 'to be' appears as a free morpheme in the present tense in formal Persian, it is realized as an enclitic, inflected for person and number, in affirmative sentences in informal Persian (e.g. *hast-and* vs *-an* 'are'). Moreover, a significant number of verbal stems appear in different forms than the formal Persian (*mi-tavân-ad* vs *mi-tun-e* 's/he can', *mi-rav-ad* vs *mi-r-e* 's/he goes'). Note that they do not follow a regular pattern.

---
[5] Within the glosses, the following abbreviations are used: ADD: additive particle, ASP: aspect, COP: copula, DOM: differential object marker, EMP: emphatic particle, EZ: ezafe, INDF: indefinite, INF: infinitive, INTJ: interjection, NEG: negation, PL: plural, PRS: present, PST: past, RED: reduplicant, SG: singular, SBJV: subjunctive.

Pronominals in Persian have both free and clitic forms. Their clitic forms, however, exhibit an extremely different pattern of behavior in two varieties. Unlike formal Persian, where pronominal clitics are typically employed to index possessive constructions, informal Persian makes extensive use of them to mark direct objects (2a), indirect objects (2b) and objects of preposition (2c) as well ('=' denotes a clitic morpheme boundary).

(2) a. diruz (shomâ-ro) did-an=**etun**
yesterday 2PL-DOM see.PST-3PL=2PL
'They saw you (pl) yesterday.'
b. basta-ro tahvil=**esh** dâd-am
parcel-DOM deliver=3SG give.PST.1SG
'I delivered the parcel to him.'
c. az=**am**
from=1SG
'from me'

The formal differential object morpheme *ra* is represented by two other allomorphs, *-o* or *-ro* in the colloquial variety, as can be seen in (1)(2). In addition, while Persian has different ways of expressing the indefiniteness of noun phrases, it lacks a definite marker that corresponds to the English definite article *the*. However, the conversational informal Persian includes a stress-attracting definite marker *-e* (*-ye* after the vowel /i/ and *-he* after other vowels) that appears on bare nominals (*ostad-e* (استاده) 'the professor', *nax-e* (نخه) 'the string', *shirini-je* (شیرینیه) 'the pastry' and *xune-he* (خونهه) 'the house'.

Persian also possesses ezafe, lit. 'addition', marked by the unstressed morpheme *-e* (*-je* if it follows a vowel). This morpheme appears between a noun and its modifier (*N-e Mod*), and is repeated on subsequent modifiers, if they are present, except the last one (*N-e Mod₁-e Mod₂-e Mod₃*). The ezafe morpheme *-e* is not written in Persian orthography, as in *aks ghshng to* (عکس قشنگ تو) 'your pretty picture'. Recently, however, it has been occasionally written down in informal texts used in social media platforms: *aks-e ghshng-e to* (عکسه قشنگه تو) (see Kahnemuyipour, 2014; Karimi and Brame, 2012; Ghomeshi, 1997; Samiian, 1983 for various approaches to this element). Note that the definite morpheme and the ezafe morpheme are homographs in the informal variant.

Reduplication is a morphological process that repeats the morphological base fully or partially (Kiparsky, 1987; Haspelmath, 2002). Persian has been argued to have three types of reduplication, namely m-reduplication (3a), intensive reduplication (3b), and indifference reduplication (4) (Smith, 2020; Ghaniabadi et al., 2006). Nouns, verbs, adjectives, adverbs, and interjections productively reduplicate to express emphasis, severity, density, and semantic extension in informal Persian, contrary to the formal variety.

(3) a. ketâb metâb
book RED
'book and other such things'
b. xis-e xis
wet-EZ wet
'completely wet'

(4) Speaker A: Nikân az xune raft
Nikan from home go.PST.3SG
'Nikan left home.'
Speaker B: raft ke raft
go.PST.3SG that go.PST.3SG
'I do not care that he left.'

Furthermore, interjections and discourse markers are frequently used in informal Persian. For instance, the formal adjective/adverb *âxar* 'finally/last' changes to *âxe* in the informal grammar, displaying a considerable semantic difference in this variant. This element expresses a wide range of emotions such as disapproval, annoyance, resentment, complaint, and surprise (5).

*Context: It is noon, and Nikan is still sleeping! His mom is very upset and angry about it! She shouts:*
(5) âxe tâ key mi-xâ-y be-xâb-i
INTJ till when ASP-want.PRS-2SG SBJV-sleep-2SG
'Until when do you wanna sleep?!!!'

Finally, technology has resulted in a great number of loanwords in informal Persian, especially from English. This variant productively generates complex predicates using loanwords as NVEs (6).

(6) kâment gozâshtan 'to comment'
espoyl kardan 'to spoil (a movie)'
fâlo/ânfâlo kardan 'to follow/unfollow'

### 3.1.3 Syntax

Although written Persian demonstrates a strict Subject-Object-Verb order, except for the sentential arguments of the verb which appear in post-verbal position, colloquial spoken Persian exhibits a fair amount of flexibility in word order. Scrambling is a syntactic property that allows phrasal categories to appear in different positions within the clause. All phrasal arguments and adjuncts are subject to scrambling in Persian. Scrambling in this language represents discourse functions such as focus and topic, depending on the stress the scrambled element carries (Karimi, 2005). Persian, interestingly, also allows for long-distance scrambling of multiple elements. The embedded subject, specific direct object, and indirect object have moved into the matrix clause in (7a). Likewise, unlike the formal variant, the adverbial nonverbal element of a complex predicate may follow the light verb in informal grammar (7b).

(7) a. gol-o$_j$ Nikân$_i$ be Jinâ$_k$ fek mi-kon-am
flower-DOM Nikan to Jina thought ASP-do-1SG
(ke) e$_i$ e$_j$ e$_k$ dâd
that give.PST.3SG
'As for the flower, as for Nikan, it was to JINA that I think (he) gave (it).' (*e* shows the original position of the elements)
b. sedâ=sh-o e$_i$ bord bâlâ$_i$
voice=3SG-DOM carry.PST.3SG up
'He raised his voice.'

Furthermore, the preposition is typically elided in locative and temporal prepositional phrases in the informal variety. The preposition *be* 'to' in the formal sentence in (8a) is deleted, and the object of preposition moves to a post-verbal position (8b).

(8) a. (man) fardâ  [be sinamâ]  mi-rav-am
    I    tomorrow  to movies   ASP-go.PRS-1SG
    'I go to the movies tomorrow.'
   b. (man) fardâ  e_i  mi-r-am  [be sinamâ]_i

Serial verb construction (SVC) is a syntactic phenomenon in which a monoclausal construction contains multiple independent lexical verbs without any linking element or predicate-argument relation, typically encoding a single event (Haspelmath, 2016). Although there has been only one study on SVCs in Persian (Nematollahi, 2015), the verbal structure in (9) appears to be a SVC, used only in the informal variant of the language. The verbs *gereft-am* and *xâbid-am* occur consecutively, and encode one event, without any coordinating or subordinating conjunction in between to indicate that one is coordinated or subordinate to the other.

(9) gereft-am      xâbid-am
    take.PST-1SG   sleep.PST-1SG
    'I slept.'

Another property of informal Persian has to do with the way the particle *ham* is employed. This formal particle follows its semantic associate as a free morpheme, and has been argued to be an additive discourse particle in Persian literature (Ghomeshi, 2020; Stilo, 2004). In the informal variety, it is typically realized as a bound morpheme *-am* (10) and is more prevalent.

(10) a. ânhâ ham  parastâr  hast-and          (formal)
        they ADD  nurse     COP.PRS-1SG
        'They are nurses, too. (besides Jina being a nurse)'

    b. unâ-m       parastâr-an              (informal)
       they-ADD    nurse-COP.PRS.1SG

Similarly, the particle *ke* can function as a discourse particle in mono-clausal sentences with an emphatic interpretation in the informal variant (11) (Ghomeshi, 2020; Lazard, 1992; Windfuhr, 1979).

(11) Simâ ke  mi-r-e          ke
     Sima EMP ASP-go.PRS-3SG  EMP
     'SIMA is going (isn't she?)'(Ghomeshi, 2020: 66, Example 15b)

The other function of this particle, only found in the informal variety, is conveying an attitude of indifference, defiance, or unbelief when it appears between a verb and its reduplicant, derived by the reduplication process, shown earlier in example (4).

Another aspect of informal Persian syntax is the deletion of the conditional conjunction *age* 'if'. This conjunction is typically present in the formal variety. The formal sentence in (12a) may or may not be realized with *age* (12b), yet it is interpreted as a conditional statement.

(12) a. agar ghazâ=yash râ   kâmel  be-xor-ad,
        if   food=3SG   DOM  complete SBJV-eat.PRS-3SG
        u    râ   be pârk  mi-bar-am
        him  DOM  to park  ASP-take.PRS-1SG
        'If he finishes his food, I'll take him to the park.'

    b. (age) ghazâ=sh-o      kâmel    bo-xor-e,
       if    food=3SG-DOM   complete  SBJV-eat.PRS-3SG
       mi-bar-am=esh            pârk
       ASP-take.PRS-1SG=him    park

Furthermore, in many cases, the coordinating conjunction *va* 'and' is dropped in the informal grammar. Also, speakers tend to use commas in writing to separate several clauses that are not necessarily coordinated.

A final distinctive aspect of informal Persian syntax relevant to this study is the usage of elliptical constructions. Ellipsis is a phenomenon in which one or more elements of a sentence are missing, yet the sentence can be fully interpreted due to the linguistic context shared by the speaker and hearer. Although ellipsis is rarely used in the formal variety, it is prevalent in the informal one (13).

(13) Jinâ ketâb xarid        vali Nikân daftar
     Jina book  buy.PST.3SG  but  Nikan notebook
     'Jina bought a book but Nikan a notebook.'

In summary, we demonstrated that the two grammars differ on a variety of levels, from phonology to morphology and syntactic structure. With the informal variety becoming dominant in writing, this variety calls for special attention in the natural language processing tasks.

### 3.2 Data Collection and Annotation Process

We crawled informal texts from open access Persian blogs. We did, however, anonymize all of the data by not including any user/author information. Furthermore, we randomized sentences in order not to have an entire blog post as a contiguous piece of text. After eliminating sentences that were suspicious in length or content (e.g., a sequence of digits or special characters), we ended up with a total of 3000 sentences. In the next step, the raw sentences were fed to the Stanza parser, a fully neural pipeline for text analysis, including tokenization, multiword token expansion, lemmatization, part-of-speech and morphological feature tagging, and dependency parsing (Qi et al., 2020). The Stanza parsing system was trained on the Uppsala UDT as it was the only publicly available universal dependency treebank for Persian at the time. In the final step, the first author manually corrected the errors using her intuition as a native-speaker as well as her linguistic knowledge. To speed up the process, we also trained a PhD candidate in linguistics who was a native speaker of Persian to help with the annotation. ConlluEditor (Heinecke, 2019) was used to visualize the dependency trees in order to expedite the process and minimize visual mistakes. 90% of the sentences were annotated once, followed by a manual check. However, to estimate the agreement between the two annotators, 10% of the sentences were annotated twice (see 3.4). After finalizing the annotation of all raw sentences, we systematically searched the treebank to find the potentially inaccurate or inconsistent annotations.

Our annotation guidelines were based on the UD annotation scheme (Nivre et al., 2020), an international collaborative ongoing project that aims to develop cross-linguistically consistent treebanks available for a wide range of languages. Language-specific guidelines of the Uppsala UDT (Seraji et al., 2016) were followed as well. This corpus was the only universal dependency treebank available at the time, and our goal was to create a treebank with a consistent annotation of grammar across different domains of Persian (in this case, formal and informal) to facilitate multi-domain parser development. We did, however, make changes to the Uppsala UDT guidelines, as in the process of annotating the automatically parsed sentences of our informal treebank, we discovered that it had several linguistically grounded problematic annotations and that it did not respect the UD guidelines in many cases (see also Rasooli et al., 2020). The presence of annotation errors or inconsistencies is detrimental to the intended uses of treebanks, especially the UD treebanks, as the main goal is to develop consistent treebank annotation for various languages. Although errors might be occasionally harmless, the proliferation of errors negatively impacts both the training and evaluation of natural language processing tasks. Detecting and removing the erroneous annotation can potentially enhance the performance of the parsers and improve multilingual parser development as well as cross-lingual learning. Therefore, through data analysis, we detected all the linguistically inaccurate or UD non-compliant annotations in the Uppsala UDT, identified the patterns, and designed linguistic rules to fix the errors. We corrected the errors manually if automatic correction was not feasible.[6] The scheme was also expanded to accommodate language-specific syntactic relations that were not included in the Uppsala UDT scheme, presumably because they were not found in the formal Persian data.

### 3.3 New Dependencies

With the exception of the proper noun (PROPN) tag which is missing in the Uppsala UDT, the parts of speech tag set (16 UPOS and 26 XOPS tags) in this treebank is identical to that of Uppsala treebank. With respect to morphological features, we added the feature *Typo*, as informal texts, unlike formal texts, are prone to typos, particularly because a phoneme may be represented by different letters in this language. Hence, individuals may become perplexed when deciding on a proper grapheme. As recipients are more tolerant of spelling and grammatical errors in informal texts, writers are less likely to be strict or careful when writing comments, blogs, and other forms of online communication. This usually results in misspelled words in informal writings (شصت instead of شست when the target word is 'thumb').

Furthermore, due to the unique properties of informal Persian, we extended the dependency relations by adding five new relations, *orphan*, *discourse*, *discourse:top/foc*, *compound:redup*, and *compound:svc*, to cover elliptical constructions, discourse markers, additive and emphatic discourse particles, reduplicative words, and serial verb constructions, respectively, discussed in 3.1. Our treebank had 51, 202, 556, 73, and 42 cases of these new syntactic relations, respectively.

Due to the poorly-edited words in informal texts, we also added the dependency relation *goeswith*. Persian orthography is not consistent in that different components of compound nouns as well as pronominal clitics, copular clitics, and certain inflectional morphemes in multiword expressions can be written joined or delimited by whitespace or a zero width non-joiner character. However, these elements are usually either attached or separated by whitespace in informal texts, contrary to formal writing, which mostly utilizes a zero width non-joiner character. Normalizing the spacing inconsistencies in such cases would render the treebank insufficient to represent informal Persian. We thus did not normalize them, and if they were separated by whitespace, we treated them as distinct tokens in order to remain faithful to the poorly edited informal source texts. Instead, we utilized the relation *goeswith* to indicate that these parts should be written together as a single token according to the orthographic rules of the language. We found 297 cases of this relation in our treebank.

### 3.4 Treebank Statistics

The basic statistics of the iPerUDT is given in Table 1. The data is split into training (80%), development (10%) and test (10%) set. Multiword expressions account for 7.44% of all tokens in iPerUDT, whereas they account for only 0.84% and 1.41% of tokens in the Uppsala UDT and PerUDT, respectively. This provides empirical evidence in favor of our claim in 3.1.2 that pronominal clitics, copula clitics, and the additive discourse bound morpheme *-am*, which create multiword expressions, are frequently used in informal Persian, contrary to formal Persian.

| Number of Sentences | 3000 |
|---|---|
| Number of tokens | 54,904 |
| Average Sentence Length | 18.3 |
| Number of unique tokens | 10889 |
| Number of multiword expressions | 4091 |

Table 1: Basic statistics of iPerUDT.

We also computed the Inter-Annotator Agreement, using Cohen's Kappa coefficient (Table 2). This statistic measures the pairwise reliability between two annotators for categorical items (Cohen, 1960). The high Kappa values can be attributed to the fact that the two annotators were jointly trained on the first 200 sentences, and for each exception encountered, the annotation guidelines were updated with clear instructions that were followed throughout the process.

| Category | Level of agreement | $\mathcal{K}$ value |
|---|---|---|
| Lemma | Almost perfect | 0.9798 |
| UPOS | " | 0.9808 |
| XPOS | " | 0.9851 |
| Head | " | 0.9859 |
| Dependency relation | " | 0.9717 |

Table 2: Inter-annotator agreement calculated for five annotation tasks in iPerUDT using Cohen's Kappa coefficient ($\mathcal{K}$).

## 4. Experiments and Results

Stanza (Qi et al., 2020) and PaT (Vacareanu et al., 2020) parsers were trained using modified training sets of formal Persian treebanks, Uppsala UDT and PerUDT, which were comparable to the iPerUDT annotation scheme. We present results only for the dev sets in order to avoid revealing details about the test sets. First, we evaluated both parsers on their own dev sets to obtain a baseline for in-domain parsing. The same trained models were then evaluated on the iPerUDT dev set (out-domain) to examine if they exhibit performance drop due to domain shift. The unlabeled attachment scores (UAS) and labeled attachment scores (LAS) are given in Table 3. The evaluation script from the 2018 UD Shared Task (Zeman et al., 2018) was used.

---

[6] We do not discuss the annotation errors because of space constraints. The details and the modified version of the Uppsala Persian Universal Dependency treebank are publicly available: https://github.com/royakabiri/modified_seraji

Regardless of the treebank it is trained on, Stanza outperforms PaT both in UAS (91.87 vs 86.32 Uppsala UDT; 94.16 vs 91.3 PerUDT) and LAS (89.65 vs 83.14 Uppsala UDT; 92.68 vs 89.14 PerUDT) when evaluated in-domain. The architectural differences between the two models can account for the higher performance of Stanza. Using a quadratic algorithm, Stanza generates probability scores for all the possible combinations of dependents and heads in a sentence, i.e. for every token, all the other tokens are considered potential heads. PaT, on the other hand, is linear and predicts the position of the head for a given token directly from the token, without taking into account the head or its vector representation. For example, for the English sentence 'Jina reads books', the head of the token 'Jina' is predicted to be +1 (because 'reads' is one position to the right) by using only the information that the BiLSTM produced for 'Jina'. Therefore, it requires a huge amount of data to converge, as LSTMs generalize better when trained on large datasets. Stanza, on the other hand, takes into account both the token 'Jina' and the head 'reads' as well as their embeddings in the prediction.

|  |  | Uppsala UDT | | PerUDT | |
| --- | --- | --- | --- | --- | --- |
| Model |  | UAS | LAS | UAS | LAS |
| Stanza | ID | 91.87 | 89.65 | 94.16 | 92.68 |
|  | OD | 83.91 | 77.33 | 83.19 | 77.63 |
| PaT | ID | 86.32 | 83.14 | 91.3 | 89.14 |
|  | OD | 80.66 | 72.7 | 80.55 | 73.76 |

Table 3: Accuracy comparison between Stanza and PaT trained on Uppsala UDT and PerUDT. In-domain (ID) indicates that the model was evaluated on the dev set of the formal treebank it was trained on. Out-domain (OD) indicates that the model was evaluated on the dev set of iPerUDT. All scores reported are F1.

Theoretically, LSTMs are designed to capture all the information, including long-distance dependencies, in a sequence (Hochreiter and Schmidhuber, 1997). They may do so in English, where there is an enormous amount of training data available. They do not, however, in several other languages for which there is not much data. Yet, even in languages like English, they perform well in nearby contexts but degrade over long distances (Khandelwal et al., 2018; Adi et al., 2017). Our in-domain evaluation results on the two parsers clearly show that it is more beneficial to employ both dependent and head embeddings when there is a limited amount of data. Scrambling, a syntactic property found in many languages, including Persian, could be another factor contributing to PaT's lower performance. This feature impacts the generalization process in this model, making it more challenging since more data is required to statistically understand where different grammatical elements are located within a sentence structure. It must rely solely on dependent information to learn the position of each element. This problem, however, is accounted for by the architecture of Stanza, where the head embedding is directly incorporated into the prediction time. Thus, it makes no difference where the head appears in the sentence because the model already has access to both the dependent and head vector representations.

Unsurprisingly, both parsers achieve a better performance when trained on PerUDT than Uppsala UDT, due to the significantly bigger size of the former (29,107 vs 5997 sentences). More training data generally allows the learning algorithm to better understand the underlying input-output mapping and generalize the patterns, resulting in a higher performing model. Nonetheless, PaT exhibits a higher performance gain (+4.98 UAS; +6.00 LAS) than Stanza (+2.29 UAS; +3.00 LAS). This supports our claim that PaT requires significantly more data to generalize than Stanza does. Therefore, when trained on the larger dataset, PaT improves substantially while Stanza improves marginally.

The performance of both Stanza and PaT parsers, trained on formal training data, substantially decreases when evaluated on informal dev data. The performance of the Stanza model trained on the Uppsala UDT has a reduction of 7.96 UAS and 12.32 LAS. Similarly, when trained on PerUDT, the Stanza model in-domain performance of 94.16 UAS and 92.68 LAS decreases to 83.19 UAS and 77.63 LAS, respectively, showing a decline of 10.97 UAS and 15.05 LAS. Interestingly, Stanza achieves almost identical accuracy scores for UAS (83.91 vs 83.19) and LAS (77.33 vs 77.63), independent of which treebank it is trained on, despite PerUDT having far more training data than the Uppsala UDT. The PaT parser trained on Uppsala treebank demonstrates a performance drop of 5.66 UAS and 10.44 LAS. The performance of this parser trained on PerUDT also drops by 10.75 UAS and 15.38 LAS. Like Stanza, PaT obtains nearly the same UAS (80.66 vs 80.55) and LAS (72.7 vs 73.76) performance when trained on two different formal treebanks. This is in contrast to the scenario in which the two parsers' in-domain performance was improved when they were trained on the larger treebank. This disparity can be explained by the significant distinctions between the formal and informal Persian. Because the formal data does not have enough variation to capture critical patterns in the informal data and there exists a domain mismatch between the training and test data, adding more training data does not lead to a better predictive performance of the parsing models. Therefore, Stanza outperforms PaT in-domain whereas they both struggle out-of-domain.

Table 4 shows the tokens with the most errors when the parsers were evaluated on the informal data. The erroneous tokens are mostly associated with the distinctive features of informal Persian, discussed in 3.1. Informal forms of differential object marking (*-ro*, *-o*), free forms of pronominals (*man*, *mâ*) functioning as nominal subject or object, pronominal clitics (*-esh*, *-am*, *-eshun*, *-emun*), free and clitic forms of copulas (*-e*, *ast*, *hast*, *bud*), interjections (*kâsh*) and discourse elements (*ham*, *ke*) exhibit numerous errors. Moreover, homographs such as *to/tu* (تو) which serves as the 2SG pronoun 'you' and the informal preposition 'in'; *va/-o* (و), which functions as the coordinating conjunction 'and', and the informal differential object marking; and *ru/ro* (رو) which can be the noun 'face', the preposition 'on', and the informal object marker, as well as homonyms such as *ham* (هم), functioning both as the reciprocal pronoun 'each other' and the discourse marker, are among the tokens which yield the most errors. Similarly, when the informal form of the verb resembles the corresponding infinitive form (*kard-an* 'do.PST-3PL'/'do.PST-INF'), parsing algorithms are faced with a greater challenge. Finally, several lexical items such as *xeyli* are quite common in different contexts in the informal variety, unlike the formal variant, which may lead to more inaccurate outputs (compare percentage of this adverb in three treebanks: iPerUDT, 5.93%; Uppsala UDT, 2.28%; PerUDT, 0.92%). In summary, although both parser perform quite well on identifying the formal properties of the language, they do poorly on the informal properties.

| POS | Token | POS | Token |
|---|---|---|---|
| CCONJ | *va* 'and' (و) | INTJ | *kâsh* 'INTJ' (کاش) |
| PART | *-ro* 'differential obj marker' (رو) | AUX | *-e* 'COP.PRS.3SG' (ه) |
| PART | *-o* 'differential obj marker' (و) | AUX | *ast* 'COP.PRS.3SG' (است) |
| PRON | *-esh* 'her/him/his/it/its' (ش) | AUX | *hast* 'COP.PRS.3SG' (هست) |
| PRON | *-am* 'my' (م) | AUX | *bud* 'COP.PST.3SG' (بود) |
| PRON | *-eshun* 'them/their' (شون) | NOUN | *kard-an* 'do.PST-INF' (کردن) |
| PRON | *-emun* 'us/our'' (مون) | VERB | *kard-an* 'do.PST-3PL' (کردن) |
| PRON | *man* 'I' (من) | ADV | *xeyli* 'very' (خیلی) |
| PRON | *mâ* 'we' (ما) | ADV | *cherâ* 'why' (چرا) |
| PRON | *shomâ* 'you (pl)' (شما) | CCONJ | *yâ* 'or' (یا) |
| PRON | *to* 'you (sg)' (تو) | DET | *in* 'this' (این) |
| ADP | *tu* 'in' (تو) | NOUN | *ru* 'face' (رو) |
| PRON | *ham* 'each other' (هم) | ADP | *ru* 'on' (رو) |
| PART | *ham* 'discourse particle' (هم) | PART | *ke* 'discourse particle' (که) |

Table 4: Tokens that yielded the most errors when parsers were evaluated on the informal dev set.

We also computed the performance of Stanza and PaT over binned distance between the dependent and head. The results are shown in Table 5. When both parsers are evaluated in-domain, their performance degrades to some extent as the distance between the dependent and head increases. In general, Stanza consistently outperforms PaT across all the binned head distances. However, the performance difference between the two diminishes when both models are trained on the PerUDT. This is in line with our discussion of the PaT model's architecture earlier. PaT does not converge easily because the farther the head is, the less information is maintained in the dependent about the head. However, Stanza addresses the long-distance problem by directly integrating the position and information of the head into the prediction. When the parsers are evaluated out-of-domain, the performance of both drops substantially.

In addition, we obtained the accuracy scores for various dependency relations. The results show that the relations corresponding to informal properties were most affected by the domain shift. A performance reduction of at least 10.00 was observed in 18 dependency relations (with a frequency greater than 10) when evaluated on out-of-domain data. We discuss the performance loss in some of the dependency relations in what follows. The performance of core arguments, including the nominal subject (*nsubj*) and direct object (*obj*), decreases by a maximum of 19.58 and a minimum of 10.9, depending on the parser and the formal treebank it was trained on. This can be explained by the fact that although formal Persian demonstrates a strict SOV order, informal Persian exhibits a fair amount of flexibility in word order. Moreover, because Persian is a null-subject language, the subject may not be phonologically realized in the clause, particularly in the informal data. A non-specific object without the differential object marking may instead appear clause-initially, tricking the models into labeling them as subjects. Finally, unlike formal Persian, direct objects are also realized by pronominal clitics in the informal variety. All these unique properties of informal Persian make it more challenging for parsing algorithms to distinguish between the subject and object, resulting in more errors in the corresponding dependency relations.

|  | Stanza | | | | PaT | | | |
|---|---|---|---|---|---|---|---|---|
|  | Uppsala UDT | | PerUDT | | Uppsala UDT | | PerUDT | |
|  | ID | OD | ID | OD | ID | OD | ID | OD |
| to root | 94.32 | 83.00 | 96.71 | 91.33 | 91.15 | 83.67 | 95.68 | 87.00 |
| 1 | 97.43 | 92.28 | 97.87 | 91.22 | 96.41 | 91.25 | 97.48 | 90.46 |
| 2 | 90.58 | 84.09 | 93.68 | 84.58 | 86.71 | 80.40 | 91.51 | 81.93 |
| 3-5 | 85.72 | 82.94 | 91.88 | 83.50 | 81.59 | 80.14 | 89.36 | 81.68 |
| 6-10 | 88.76 | 79.53 | 93.28 | 78.44 | 84.99 | 80.46 | 90.82 | 75.89 |
| 11-15 | 90.56 | 79.26 | 89.63 | 67.32 | 84.61 | 71.83 | 86.23 | 67.57 |
| 16-21 | 88.27 | 84.21 | 88.84 | 76.93 | 78.94 | 68.75 | 82.96 | 63.92 |
| 21- … | 89.43 | 74.19 | 86.87 | 48.48 | 83.37 | 65.46 | 82.05 | 36.67 |

Table 5. Accuracy comparison of in-domain (ID) and out-domain (OD) performance of Stanza and PaT over binned head distances. All scores reported are F1.

The performance drop scores obtained for copula (*cop*) and auxiliary (*aux*) dependency relations are at most 22.86, and at least 10.75. Recall that the copula *budan* 'to be' appears as a free morpheme in the formal variety but as an enclitic in the informal variety. When it is realized as an enclitic, it is homograph with some other widely-used elements in the language (e.g. 1SG copula clitic, 1SG pronominal clitic, and 1SG verbal agreement suffix all appear as *-am*). These challenges account for the considerable decline in the relevant dependency relations performance on informal data.

Given the dramatic change in informal Persian verbal construction, we can explain why the *compound:lvc* and *compound:prt* labels show a lower performance when evaluated on the informal texts. For the former, the performance drop ranges from 10.21 to 16.89, and for the latter, it ranges from 21.47 to 58.15. First, the adverbial nonverbal element of a complex predicate may follow the light verb in informal grammar (see 3.1.3). Second, the agreement inflections appear in slightly different forms (see 3.1.2), resulting in homographs: the 3PL agreement suffix and the infinitive marker both appear as *-an* in the informal variant. Since infinitives usually function as nouns, it makes it challenging for the algorithms to differentiate between the verbal constructions and the nominal phrases. Furthermore, complex predicates with English loanwords NVEs that were unseen in the training data contributed to the poor performance of the models.

For all types of subordinate clauses, including complement clause (*ccomp*), adverbial clause (*advcl*), adnominal clause (*acl*) and relative clause (*acl:relcl*), the performance is reduced to some extent. The performance drop can be attributed to the fact that the complementize *ke* 'that' optionally introduces complement clauses in Persian and it is rarely used in the informal variety. In addition, the complementizer *ke* is homograph with the discourse-functioning particle *ke* in informal Persian. In general, due to word order flexibility, different word forms, and the presence of new lexical items, parsing algorithms have a harder time identifying subordinate clauses in this variant. The performance drop of *advcl* is also consistent with our descriptive analysis of informal Persian syntax (example 12), where we argued that the conditional conjunction *age* 'if' is usually elided in this variety.

The maximum performance drop scores for fixed multiword expressions (*fixed*) and flat multiword expressions (*flat*) are 56.13 and 65.02, and the minimum scores are 18.14 and 29.66, respectively. In contrast to formal Persian, the informal variety features a large number of multiword expressions, as indicated in 3.4. Specifically, fixed grammaticized expressions such as *dar-hâl-i-ke* (lit. in-mood-INDF-that) 'while' or *bâ-in-ke* (lit. with-this-that) 'although' can be written attached, by half-space or whitespace in this variety. Such expressions were not normalized. However, they were tokenized as distinct tokens (regardless of how they appeared in source texts), linked by the dependency relation *fixed*. Because of the orthographic inconsistencies in informal writing, the parsers trained on formal data are confronted with such unknown tokens and structures, resulting in poor predictions in the corresponding dependency relations.

The performance on the *parataxis* relation reduces by 85.71 and 84.00, respectively. The substantial performance drop can be explained by the fact that the informal treebank contains a higher percentage of *parataxis* relation (compare 0.45% iPerUDT with 0.14% Uppsala UDT and 0.016% PerUDT), supporting our claim in 3.1.3 that informal Persian writing typically elides the coordinating conjunction *va* 'and' and uses commas to separate different clauses. Finally, different syntactic structures between the two varieties can account for the low performance of root (maximum and minimum performance drop scores are 11.32 and 10.35).

We then trained the two parsers on the training data of our informal treebank and evaluated them on its dev set (Table 6).

|       | iPerUDT |       |
|-------|---------|-------|
| Model | UAS     | LAS   |
| Stanza| 90.89   | 86.83 |
| PaT   | 79.82   | 74.09 |

Table 6: Accuracy scores for Stanza and PaT trained on iPerUDT and evaluated on its dev set. All scores reported are F1.

Although the size of our treebank is much smaller than the two other formal treebanks, Stanza achieves a better performance score than when it is trained on the formal treebanks (iPerUDT: 90.89 UAS and 86.83 LAS; Uppsala UD: 83.91 UAS and 77.33 LAS; PerUDT: 83.19 UAS and 77.63 LAS (see Table 3)). PaT, however, obtains a lower UAS score and only a slightly higher LAS score compared to a scenario where the model is trained on the formal data (iPerUDT: 79.82 UAS and 74.09 LAS, Uppsala UD: 80.66 UAS and 72.7 LAS, PerUDT: 80.55 UAS and 73.76 LAS). These results support our previous discussion of the parsers' architectures, which indicated that PaT requires a large amount of data to learn the patterns, whereas Stanza does not. Hence, it does not perform well in-domain where there is a limited amount of training data.

## 5. Conclusion

In this paper, we showed that the grammatical distinctions between the formal and informal variants of Persian are often as stark as those between two distinct dialects of a language. We then presented the Informal Persian Universal Dependency Treebank, a new treebank created in this effort, and provided empirical evidence for the need of this treebank that focuses explicitly on informal language. Our experiments revealed that the dependency parsers, primarily trained on formal data, experience a substantial performance loss when evaluated on informal text. Specifically, it was shown through examples that the majority of erroneous tokens and dependency relations reflect the distinctive characteristics of informal grammar. It must be noted that the performance of other Stanza modules (Qi et al., 2020) including tokenizer, lemmatizer, and POS tagger, also decreased considerably when evaluated on the informal text. However, we leave further research into these tasks to the future. This study may highlight the importance of informal variants of languages and inspire researchers to further investigate such varieties.

## 6. Acknowledgments

We would like to thank Roshan Cultural Heritage Institute and Dr. Malakeh Taleghani Graduate Fellowship in Iranian Studies for their support and financial contribution. We are also grateful to Farzaneh Bakhtiary for her help with the annotation.

Universal Dependencies. In *Proceedings of the CoNLL 2018 Shared Task: Multilingual Parsing from Raw Text to Universal Dependencies*, 1–21, Brussels, Belgium, Association for Computational Linguistics.

## 8. Language Resource References